\pdfoutput=1

\documentclass[10pt,twocolumn,letterpaper]{article}

\usepackage{cvpr}              

\usepackage[dvipdfm]{graphicx}
\usepackage{bmpsize}
\usepackage{amsmath}
\usepackage{amssymb}
\usepackage{booktabs}
\usepackage{mathtools}
\usepackage{multirow}
\usepackage{soul}

\usepackage[pagebackref,breaklinks,colorlinks]{hyperref}

\usepackage[capitalize]{cleveref}
\crefname{section}{Sec.}{Secs.}
\Crefname{section}{Section}{Sections}
\Crefname{table}{Table}{Tables}
\crefname{table}{Tab.}{Tabs.}

\usepackage{algpseudocode}
\usepackage{algorithm}
\usepackage{breqn}

\def\cvprPaperID{7}
\def\confName{CVPR}
\def\confYear{2022}

\newif\ifworkshop
\newif\ifmain
\maintrue

\usepackage{overpic}
\usepackage{enumitem} 
\usepackage{overpic} 
\usepackage{color}

\usepackage[dvipdfm]{graphicx}
\usepackage{bmpsize}
\usepackage{adjustbox}
\usepackage{multicol, multirow}
\usepackage{subcaption}

\definecolor{turquoise}{cmyk}{0.65,0,0.1,0.3}
\definecolor{purple}{rgb}{0.65,0,0.65}
\definecolor{dark_green}{rgb}{0, 0.5, 0}
\definecolor{orange}{rgb}{0.8, 0.6, 0.2}
\definecolor{red}{rgb}{0.8, 0.2, 0.2}
\definecolor{darkred}{rgb}{0.6, 0.1, 0.05}
\definecolor{blueish}{rgb}{0.0, 0.3, .6}
\definecolor{light_gray}{rgb}{0.7, 0.7, .7}
\definecolor{pink}{rgb}{1, 0, 1}
\definecolor{greyblue}{rgb}{0.25, 0.25, 1}

\DeclareMathOperator*{\argmin}{arg\,min}
\DeclareMathOperator*{\argmax}{arg\,max}

\newcommand{\Figure}[1]{Figure~\ref{fig:#1}}

\newcommand{\Table}[1]{Table~\ref{tab:#1}}

\newcommand{\Eq}[1]{Eq.~\ref{eq:#1}}

\usepackage{blindtext}

\renewcommand{\paragraph}[1]{\vspace{1em}\noindent\textbf{#1}.}

\newcommand{\myparagraph}[1]{\smallskip\noindent\textbf{#1.}}

\def\A{\mathcal{A}}
\def\E{\mathcal{E}}
\def\T{\mathcal{T}}

\begin{document}

\title{Effectiveness of Adversarial Training for Video Action Recognition}
\title{Adversarial Training for Video Classification: Empirical Analysis and New Algorithms}
\title{Optimal Attacks and Defenses for Video Classification}
\title{Analysis and Extensions of Adversarial Training for Video Classification}

\author{Kaleab A. Kinfu and Ren\'e Vidal\\
Mathematical Institute for Data Science,
Johns Hopkins University, Baltimore, MD, USA
}
\maketitle
\begin{abstract}
Adversarial training (AT) is a simple yet effective defense against adversarial attacks to image classification systems, which is based on augmenting the training set with attacks that maximize the loss. However, the effectiveness of AT as a defense for video classification has not been thoroughly studied.  Our first contribution is to show that generating optimal attacks for video requires carefully tuning the attack parameters, especially the step size. Notably, we show that the optimal step size varies linearly with the attack budget. Our second contribution is to show that using a smaller (sub-optimal) attack budget at training time leads to a more robust performance at test time. Based on these findings, we propose three defenses against attacks with variable attack budgets. The first one, Adaptive AT, is a technique where the attack budget is drawn from a distribution that is adapted as training iterations proceed. The second, Curriculum AT, is a technique where the attack budget is increased as training iterations proceed. The third, Generative AT, further couples AT with a denoising generative adversarial network to boost robust performance. Experiments on the UCF101 dataset demonstrate that the proposed methods improve adversarial robustness against multiple attack types.
\end{abstract}

\vspace{-3mm}
\section{Introduction}
\label{sec:intro}
%


Deep neural networks (DNNs) have led to significant advances in many computer vision tasks, including image classification, object detection, and image segmentation. However, a key challenge to DNNs is their vulnerability to adversarial attacks~\cite{Szegedy2014IntriguingPO}, i.e., small perturbations to the input data that fool the classifier.
{This has motivated the development of various defense mechanisms against such attacks.}
Among them, Adversarial Training (AT)~\cite{Goodfellow2015ExplainingAH, Madry2018TowardsDL, Zhang2019TheoreticallyPT}, a technique that 
{extends
the training set with examples that maximize the loss,}
shows state-of-the-art adversarial robustness.
Nevertheless, the robust performance against adversarial attacks is still far from perfect
as there is a significant difference between clean accuracy 
and adversarial accuracy.

In this work, we are interested in studying the adversarial robustness of DNNs for classifying actions in video data. Action classification aims to predict a single action label for a video and is one of the main tasks in video understanding. Numerous DNN-based methods have been developed for this task, \eg, \cite{Feichtenhofer2016SpatiotemporalRN, Hara2018CanS3, Kataoka2020WouldMD,Carreira2017QuoVA}, which exploit the spatio-temporal information from consecutive frames in the video.  

Unlike image attacks, video attacks consider not only spatial but also temporal information. For example, flickering attacks~\cite{Pony2021OvertheAirAF} generate a flickering temporal perturbation whereas other attacks, such as the frame saliency attacks~\cite{Inkawhich2018AdversarialAF}, generate temporally sparse perturbations. Moreover, AT has been found to be difficult to successfully train a robust model on large-scale problems~\cite{Kurakin2017AdversarialML}, thus it is a natural extension to study the effectiveness of AT for videos. 

Although adversarial attacks in the case of videos are not as investigated as in the image domain, a few attack generating methods have been recently proposed~\cite{Jiang2019BlackboxAA, Wei2019SparseAP,Inkawhich2018AdversarialAF, Pony2021OvertheAirAF}. Similarly, defenses in the video domain are less explored. Xia \etal \cite{Xiao2019AdvITAF} proposed adversarial frame detection from videos leveraging temporal consistency by applying optical flow estimation to consecutive frames. However, this technique can only tell if the video is adversarial or not. Lo \etal \cite{Lo2020DefendingAM} proposed using AT with multiple independent batch normalization layers. However, the experiments are limited to weak attacks, and the robust performance against stronger attacks has not yet been reported.

\myparagraph{Paper Contributions}
In this paper, we study the effectiveness of AT for video classification. Since AT is based on generating attacks that maximize the loss during training, we begin by understanding how to generate such optimal attacks. 
\ifmain 
We observe that generating optimal attacks requires carefully tuning the attacker parameters, such as the attack budget, the step size, and the number of steps. In particular, we show that there is an optimal choice for the step size that changes linearly with the attack budget. Next, we focus on understanding whether optimal attacks at training time lead to better defenses at test time. Surprisingly, the answer to this question is no. Specifically, we show that using attacks with a smaller (sub-optimal) attack budget during training leads to a more robust performance at test time. Based on these findings, we propose three defenses, adaptive AT, curriculum AT and generative AT, which use sub-optimal attacks with an adaptive attack budget during training.
\fi
In summary, this paper makes the following contributions:
\begin{itemize}[leftmargin=*]
\setlength\itemsep{-.3em}
\item We show that generating optimal attacks is very sensitive to the choice of the step size and requires careful tuning. We also show that the optimal step size varies for different attack budgets and that their relationship is linear.
\item {Contrary to the common belief that the key to solving the min-max problem in AT is to find optimal adversarial examples~\cite{Huang2015LearningWA}, we empirically demonstrate that AT with an optimal attack does not lead to a strong defense. Rather, we show that AT with a smaller attack budget leads to a better defense. We also show that AT with a higher attack budget does not necessarily lead to a better defense.}
\item {Since AT with a single perturbation type does not often yield robustness to multiple attack types, we propose AT with an adaptive random sampling of variable attack types and budgets. Moreover, based on our finding that AT with optimal attacks fails, we propose curriculum based adversarial learning with an increasing attack budget as training proceeds. We also propose to combine adversarial AT with a GAN-based perturbation elimination network~\cite{Jin2019APEGANAP} as a pre-processor to improve robust performance.  Finally, we demonstrate the robustness~of~the proposed methods with experiments on the UCF101 dataset.}
\end{itemize}


\section{Related Work}
\label{sec:related}

\ifmain
In this section, we briefly review adversarial attack methods related to this work as well as the adversarial training technique that we will study. In addition, we briefly discuss generative defense techniques related to our method. 

\fi
\myparagraph{Adversarial Attacks}
Given an image $X$ and a classifier $F$ with parameter $\theta_F$, an additive adversarial attack is a small perturbation $\delta$ to the image $X$ which is designed to deliberately fool the classifier, i.e. $F(X+\delta;\theta_F) \neq F(X;\theta_F)$. The perturbation $\delta$ is assumed to be sufficiently small so that they are imperceptible to humans. A common assumption is that the perturbations lie within a ball $\mathcal{B}_\epsilon = \{\delta: \| \delta \|_p \leq \epsilon \}$, where $\epsilon$ denotes the attack budget and $p \in \{0,1,2,\infty\}$.


Ever since Szegedy \etal~\cite{Szegedy2014IntriguingPO} first observed that DNNs are vulnerable to adversarial attacks, several attack methods have been proposed by the research community. These attack methods can be classified in two categories: white-box and black-box attacks. White-box attacks assume access to the gradient information of the model, while in black-box attacks, the attacker does not have access to the parameters of the model. White-box attacks~\cite{Goodfellow2015ExplainingAH, Madry2018TowardsDL, Carlini2017TowardsET} basically maximize the loss value by iteratively computing a perturbation and adding it to the data, whereas black-box attacks~\cite{Zhang2020BlackBoxAO, Co2019ProceduralNA} are usually performed by using a different model and transferring adversarial examples to the target model.

In this work, we focus on white-box attacks obtained by maximizing the cross-entropy loss, i.e.
\begin{equation}
\hat{\delta} = \argmax_{\delta \in \mathcal{B}_\epsilon} \mathcal{L}_{ce}(F(X+\delta;\theta_F), y),
\end{equation}
where $y$ is the label for image $X$. Goodfellow \etal~\cite{Goodfellow2015ExplainingAH} propose the Fast Gradient Sign Method (FGSM), a single step gradient based attack, to compute the perturbation $\delta$:  
\begin{equation}
    \delta = \epsilon \text{ sign}(\nabla_\delta \mathcal{L}_{ce}(F(X;\theta_F),y)). 
\end{equation}
Madry \etal~\cite{Madry2018TowardsDL} extend this by introducing an iterative approach to finding the optimal attack known as Projected Gradient Descent (PGD), which can be formulated as:
\begin{equation}
\delta_{i+1} = \Pi_{\mathcal{B}_\epsilon} \left\{ \delta_i + \alpha \text{ sign}(\nabla_\delta \mathcal{L}_{ce}(F(X+\delta_i;\theta_F),y))\right\},
\label{eq:pgd}
\end{equation}
where $\Pi_{\mathcal{B}\epsilon}$ denotes the projection of the perturbation onto the norm ball $\mathcal{B}_{\epsilon}$, while $\alpha$ is the step size and $i$ is the current step. 
In this work, we will mostly consider PGD bounded $\ell_\infty$-based attacks and denote it as PGD$_\infty$.

\paragraph{Adversarial Attacks to Video Data}
So far, most adversarial attack studies have focused on the image domain, leaving attacks for videos less explored. However, there are few attack techniques for videos. \textit{Masked PGD attacks} apply the PGD$_\infty$ attack to a patch of a frame given a random top-left position and a patch ratio $r$. The patch height and width are computed by multiplying the frame height and width by $\sqrt{r}$. \textit{Frame Border attacks} apply the PGD$_\infty$ attack to the border of a frame instead of a random patch. The thickness of the border $b$ given a ratio $r$ is calculated as a function of the image size.
Beyond $\ell_{\infty}$ attacks, \textit{sparse adversarial attacks}~\cite{Wei2019SparseAP} fool action classification networks by using $\ell_{2,1}$ optimization to compute the perturbations. 

\textit{Frame saliency attacks}~\cite{Inkawhich2018AdversarialAF} extend iterative gradient-based attacks to prioritize which parts of the video frames should be perturbed based on saliency scores. The baseline variant is the \textit{one-shot attack}, which only involves one iteration of the attack. Here, the loss is computed with respect to each input frame. The other variant is an \textit{iterative-saliency} attack that generates adversarial examples by iteratively perturbing some frames of the video one at a time in order of decreasing saliency score until it fools the network.

\textit{Flickering attacks}~\cite{Pony2021OvertheAirAF} generate a flickering temporal perturbation that is imperceptible to human observers by adding a uniform offset to the $C$ color channels of the pixels in every frame of a video. The perturbation is computed by optimizing the following objective function:
\begin{equation}    
\hat{\delta} = \argmin_\delta \lambda \sum_j \beta_j D_j(\delta) + \frac{1}{N} \sum_{n=1}^N \mathcal{L}_{cw}(F(X+\delta; \theta_F),y),
\end{equation}
where $\mathcal{L}_{cw}$ denotes the C\&W loss~\cite{Carlini2017TowardsET}, $\lambda$ controls the importance of the adversarial attack relative to the regularization terms $D_j(\cdot)$, which control the imperceptibility of the attack. The parameter $\beta_j$ weights the relative importance of each regularization term. 
Here, the regularization terms are thickness regularization that enforces the perturbation to be small in all color channels and roughness regularization that penalizes temporal changes of the perturbation pattern.

\myparagraph{Adversarial Defenses}
The discovery of vulnerabilities in DNNs has raised security concerns and attracted significant attention from the research community. Accordingly, various defense methods have been proposed~\cite{He2021TowardsST}. Papernot \etal~\cite{Papernot2016DistillationAA} propose a distillation defense that uses knowledge from the distillation process to reduce the magnitude of network gradient which could successfully mitigate perturbations computed by FGSM.
Guo \etal~\cite{Guo2018CounteringAI} use a transformation technique that replaces input patches with clean patches to remove the perturbations. 
\ifmain
Dhillon \etal~\cite{Dhillon2018StochasticAP} propose applying dropout to each layer in order to prune a random subset of activations with a weighted distribution. 
\fi
While such methods do not require retraining the model, Athalye \etal~\cite{Athalye2018ObfuscatedGG} show that they suffer from the obfuscated gradients problem, which gives a false sense of robustness.

\begin{figure*}
\centering
\subfloat[Benign ($\epsilon=4$)]{%
  \includegraphics[width=.3\linewidth]{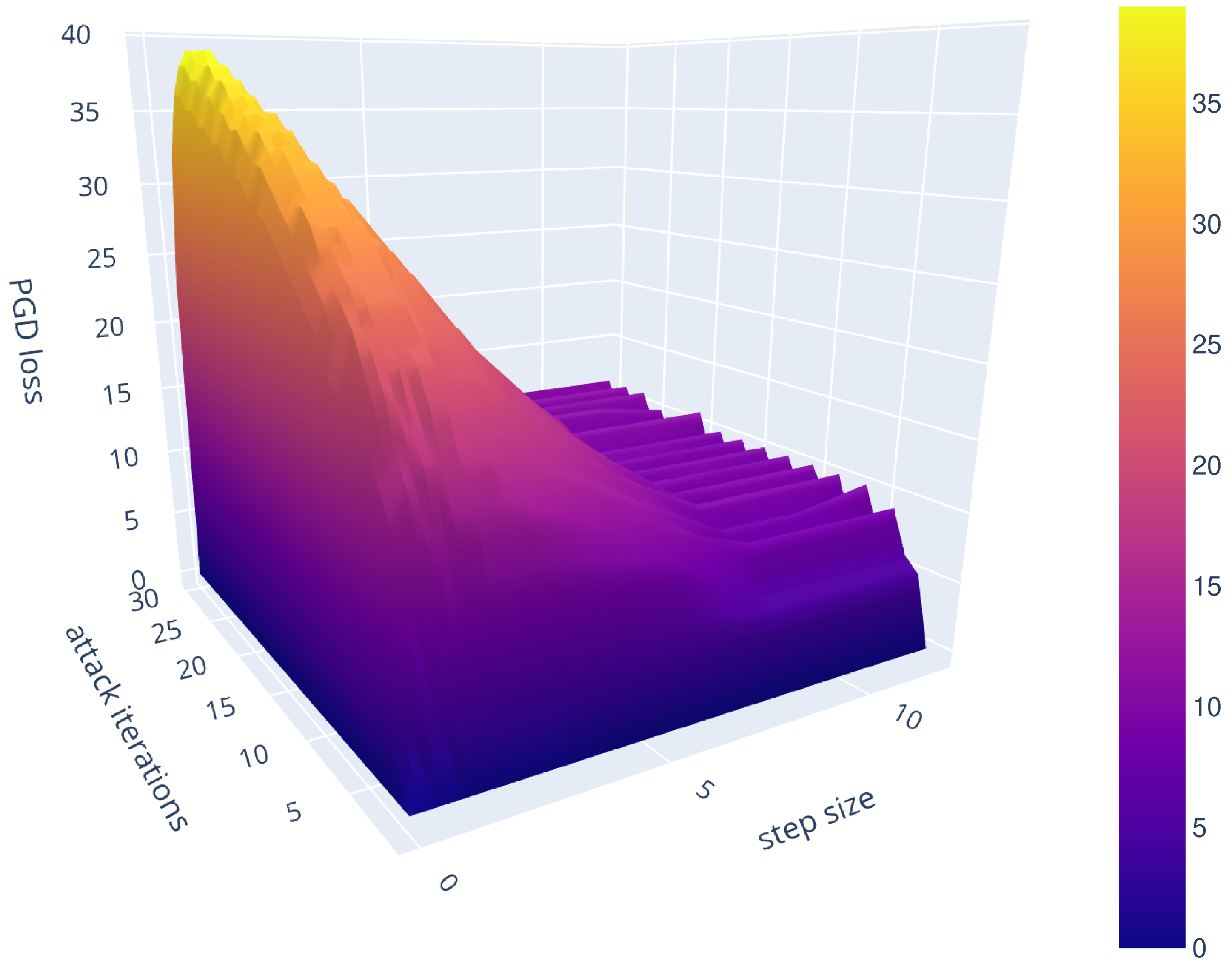}
  \label{fig:loss:eps4}%
}\qquad
\subfloat[Benign ($\epsilon=8$)]{%
  \includegraphics[width=.3\linewidth]{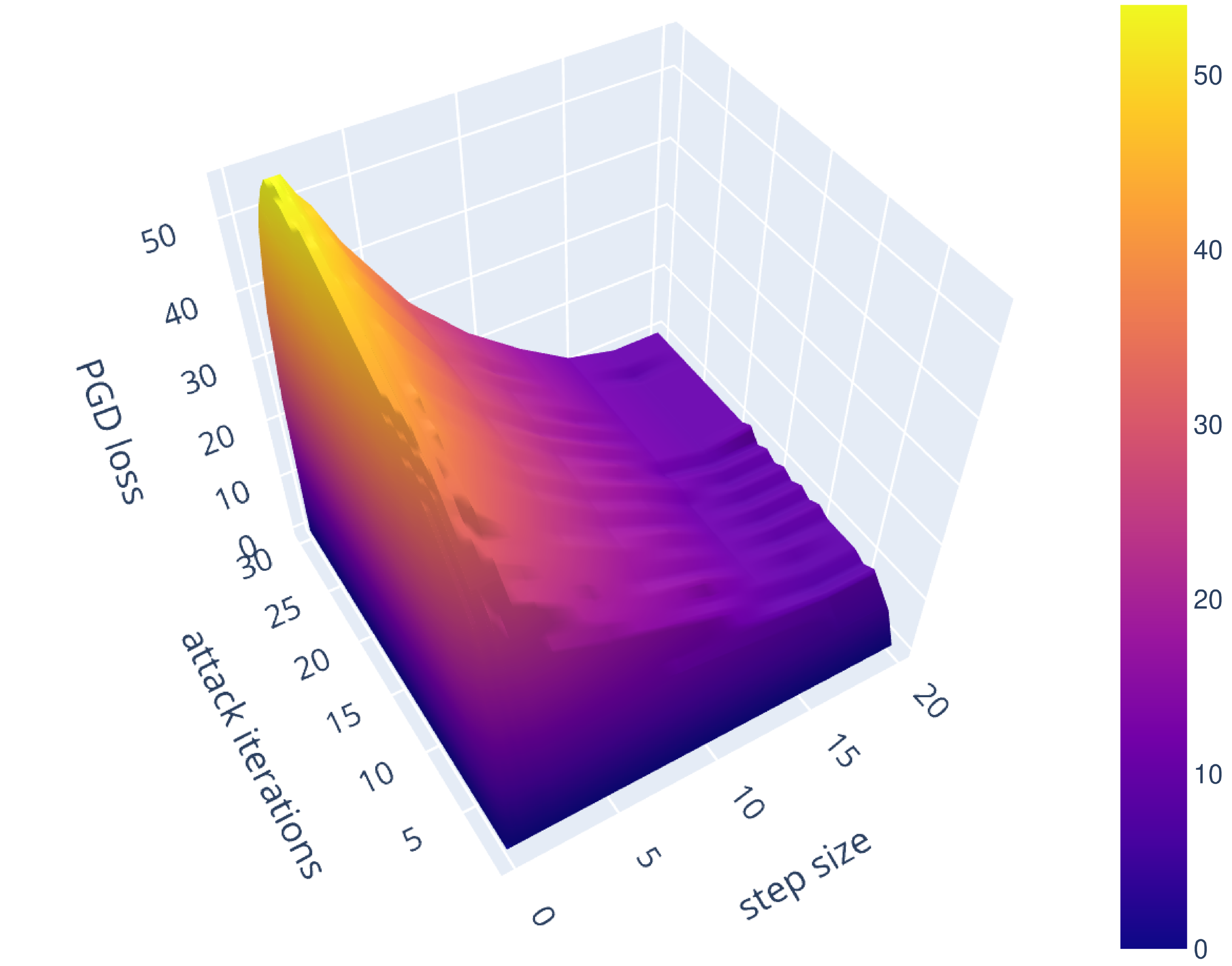}
  \label{fig:loss:eps8}%
}\qquad
\subfloat[AT ($\epsilon=8$)]{%
  \includegraphics[width=.3\linewidth]{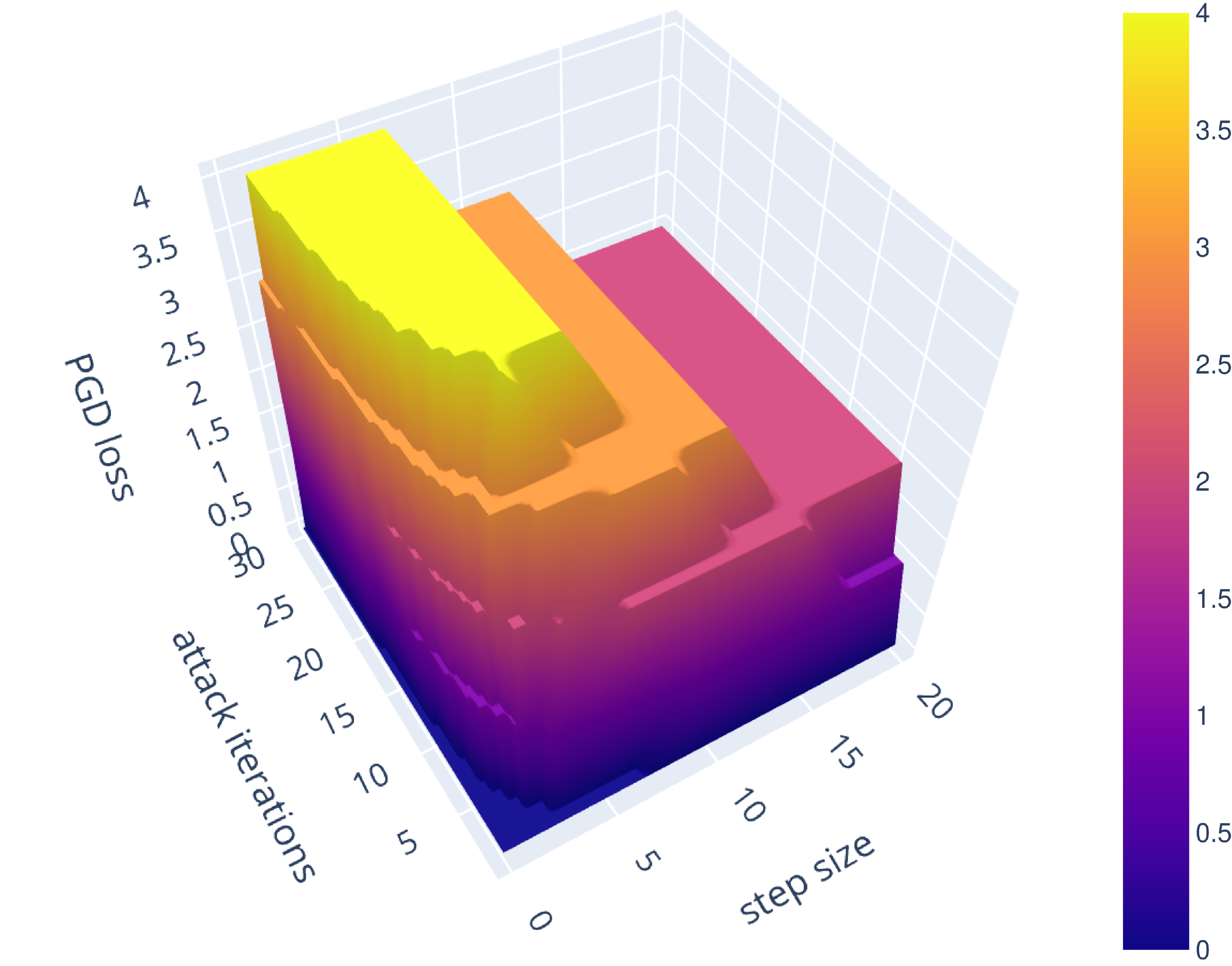}
  \label{fig:loss:eps8_at}%
}\qquad
\caption{
(a) Adversarial loss surface of a benign model trained on clean data when the input is perturbed by PGD$_\infty$ attack with a budget $\epsilon=4$, (b) Adversarial loss surface of the same benign model perturbed by PGD$_\infty$ attack with a budget $\epsilon=8$. (c) Adversarial loss surface of the same model after trained with adversarial examples for 10 epochs with a budget $\epsilon=8$. The $x$-axis is the step size $\alpha$; the $y$-axis is the number of steps $M$, and the $z$-axis is the adversarial loss value. From (a) and (b), observe that generating optimal attacks require tuning the step size in a thin region. Specifically, a small step size, i.e. below some threshold, generates a very weak attack, while increasing the step size after the optimal value also reduces the attack strength. Moreover, the adversarial loss becomes a flat when $\alpha \geq 2\epsilon$. From (b) and (c) we can see that the adversarial loss decreases significantly and the step size region that can generate the optimal attack becomes wider.
}
\label{fig:loss_plot}
\end{figure*} 

Adversarial training (AT)
\cite{Szegedy2014IntriguingPO, Goodfellow2015ExplainingAH} is based on solving a min-max optimization to achieve robustness via retraining the model with adversarial examples. 
Shaham \etal~\cite{Shaham2018UnderstandingAT} formalize AT as the following robust optimization problem: 
\begin{equation}
    \min_{\theta_F} \mathop{\mathbb{E}}_{(X,y) \sim \mathcal{D}} \left[ \max_{\delta \in \mathcal{B}_\epsilon} \mathcal{L}_{ce} (F(X+\delta;\theta_F), y) \right],
    \label{eq:at_basic}
\end{equation}
where $(X,y)$ is the training data sampled from the data distribution $\mathcal{D}$. The inner maximization problem aims to generate an optimal adversarial example for each $(X,y)$, while the outer minimization problem trains the model to be robust to such adversarial examples. 

AT using the PGD attack to solve the inner maximization problem is a widely used technique that yields state-of-the-art performance in many image-based tasks~\cite{Tramr2018EnsembleAT}. 
However,
Kurakin \etal~\cite{Kurakin2017AdversarialML} show that AT is difficult at ImageNet scale and Sharma and Chen~\cite{Sharma2018AttackingTM} show that training using $\ell_\infty$ adversarial examples yields limited robustness.

\myparagraph{Generative Adversarial Defenses}
Song \etal~\cite{Song2018PixelDefendLG} empirically show that most adversarial examples lie in low probability regions of the training data distribution, thus proposing a generative method that tries to project adversarial examples back onto the clean data manifold prior to classification. This technique is based on the idea of purification and it uses PixelCNN~\cite{Oord2016PixelRN} to approximately find the highest probability example constrained to be within an $\epsilon$-ball of a perturbed image. Similarly, Samangouei \etal~\cite{Samangouei2018DefenseGANPC} propose to use a Generative Adversarial Network (GAN), instead of PixelCNN, to project adversarial examples back onto the manifold of the generator. On the other hand, She \etal~\cite{Jin2019APEGANAP} propose an approach called Adversarial Perturbation Elimination with GAN (APE-GAN) that allows for training a GAN to systematically clean the perturbations from the adversarial examples thus learning a manifold mapping from adversarial data to clean data. 
\ifmain
Lin \etal\cite{Lin2020DualMA} propose a method known as Dual Manifold Training that uses perturbations in the latent space of StyleGAN~\cite{Karras2019ASG} as well as in the image space in order to adversarially train a model with the aim of defending against $\ell_p$ and non-$\ell_p$ attacks. However, this requires a pre-constructed On-Manifold dataset, which is too expensive for video data.
\fi

\section{Empirical Analysis of Adversarial Training with PGD Attacks for Action Classification}

In this section, we present an empirical analysis of adversarial training (AT) with PDG attacks for action classification in the UCF101 dataset. We first show that a na\"{i}ve application of PGD to action classification fails to produce optimal attacks, i.e., attacks that maximize the loss.
\ifmain
Specifically, we show that generating optimal attacks requires carefully tuning the parameters of PGD, especially the step-size. We also show that different attack budgets require a different step size in order for PGD to find an optimal attack, and that the relationship between the optimal step size and the attack budget is linear. 
\fi
As a consequence, we obtain a practical procedure for obtaining optimal attacks for action classification via suitable tuning of the PGD parameters. Next, we show that optimal attacks during training do not lead to a strong defense. Specifically, we show that choosing an attack budget at training that is smaller than the attack budget at testing leads to a more robust performance at testing. This motivates us to propose various modifications to AT and generative defenses, which will be discussed in Sec. \ref{sec:ATextensions}.

\subsection{Optimal PGD Attacks via Parameter Tuning}

Let us first recall the definition of an optimal attack for video classification. Let $X \in  \mathbb{R}^{T\times H\times W\times C}$ denote a video with $T$ frames, where $H$ and $W$ denote, respectively, the height and width of each frame, and $C$ denotes the number of color channels. Let $y\in\{1,\dots, K\}$ denote the label for the video. In this case, the perturbation is also a video denoted as $\delta = [\delta_1, \delta_2,\dots,\delta_T] \in \mathbb{R}^{T\times W\times H \times C}$, where $\delta_t$ denotes the perturbation to the $t^{th}$ frame $x_t$ of the video $X$. 

\ifmain
Given an action classification model $F$ with parameters $\theta_F$, and an attack budget $\epsilon>0$, an optimal adversarial attack is computed by finding a perturbation $\delta$ that maximizes, \eg, the cross-entropy loss function $\mathcal{L}_{ce}$, \ie:
\begin{equation}
\hat{\delta} = \argmax_{\delta \in \mathcal{B}_\epsilon} \mathcal{L}_{ce}(F(X+\delta;\theta_F), y),
\end{equation}
where $\mathcal{B}_\epsilon = \{\delta: \| \delta \|_p \leq \epsilon \}$ is the $\ell_p$-norm ball with radius $\epsilon$, and $p \in \{0,1,2,\infty\}$. The Projected Gradient Descent (PGD) method solves the above optimization problem using the following iterative algorithm for $i=1,\dots, M$,
\begin{equation}
\label{eq:pgd}
\delta_{i+1} = \Pi_{\mathcal{B}_\epsilon} \left\{ \delta_i \!+\! \alpha \text{ sign}(\nabla_\delta \mathcal{L}_{ce}(F(X\!+\!\delta_i;\theta_F),y))\right\},\!
\end{equation}
where $\Pi_{\mathcal{B}\epsilon}$ is the projection of the perturbation onto the ball $\mathcal{B}_{\epsilon}$, $\alpha$ is the step size and $M$ is the number of iterations.
\fi
\ifmain 

\fi
In order to evaluate the effects of the attack parameters $\alpha$ and $M$ in generating an optimal perturbation for a given $\epsilon$, we visualize the adversarial loss of a benign model trained on clean data. 
\ifworkshop 

\fi
We adopt a 3D CNN architecture based on residual networks (ResNets)~\cite{He2016DeepRL}, namely 3D ResNeXt-101~\cite{Hara2018CanS3}, as a backbone for the action recognition task. We adopt the training strategy and configuration of parameters from~\cite{Kataoka2020WouldMD}. 
The network was first pre-trained on Kinetics-700~\cite{Carreira2019ASN},
a large-scale video dataset with more than 650,000 videos covering 700 categories.
This step is important as such networks require an enormous amount of data.
The Kinetics-700 pretrained network was then fine-tuned on the UCF101 dataset~\cite{Soomro2012UCF101AD}, a widely used dataset in action recognition. The UCF101 dataset consists of 13,320 videos with 101 action classes providing a large diversity of actions, variations in lighting conditions, camera motion, viewpoint, and partial occlusion. The dataset is split into 9,537 videos for training and 3,783 videos for testing. 

\Figure{loss_plot} shows the adversarial loss of the benign model for a PGD$_\infty$ attack as a function of the step size $\alpha$ and the number of iterations $M$ for a fixed value of $\epsilon = 4$ (\ref{fig:loss:eps4}) and $\epsilon=8$ (\ref{fig:loss:eps8}). As it can be seen from the plot, there is a very small range of values of $\alpha$ that will generate the optimal attack. Indeed, a step size below the threshold will generate a very weak attack, while the loss becomes flat if a step size above $2\epsilon$ is used. Note that the optimal step size changes with the attack budget, since $\hat\alpha \approx 1 $ for $\epsilon = 4$ and $\hat\alpha \approx 1.8$ for $\epsilon = 8$, hence $\alpha$ must be properly tuned. 

\begin{figure}[ht]
\centering
\includegraphics[width=.9\linewidth,clip=true,trim=0 10 20 70]{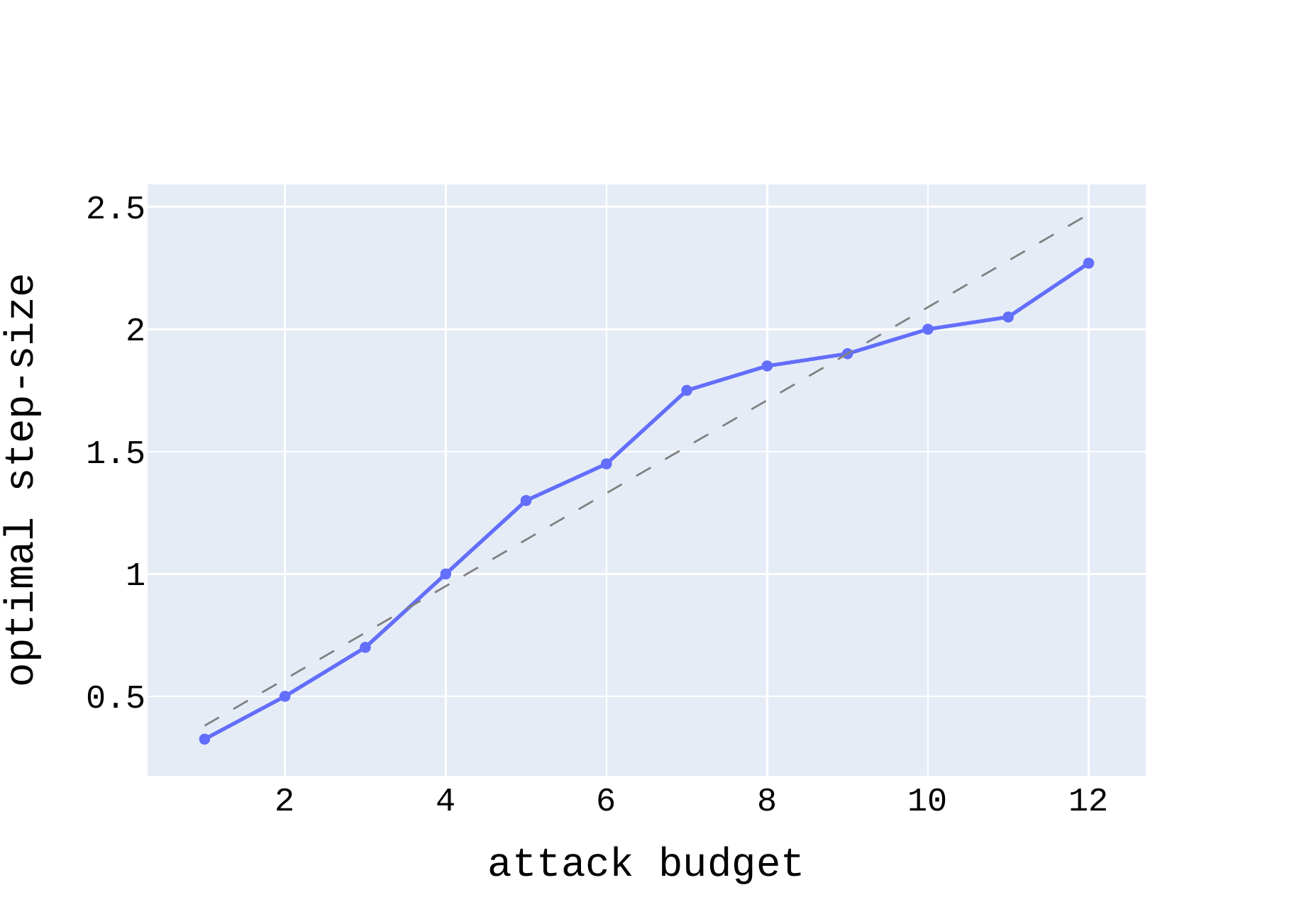}
\caption{
\textbf{Optimal step size vs attack budget:}
The relationship between the optimal step size $\hat{\alpha}$ and the attack budget $\epsilon$ is well approximated by linear regression (dash line) as $\hat{\alpha} \approx 0.2 (\epsilon + 1)$.}
\vspace{-2mm}
\label{fig:step_vs_budget}
\end{figure}

We further investigated the relationship between the optimal step size $\hat{\alpha}$ and the attack budget $\epsilon$ by computing the loss surfaces for multiple values of $\epsilon$ in the range $[1,12]$ and finding the optimal $\alpha$ for each $\epsilon$. As we can see from \Figure{step_vs_budget}, there is a linear relationship between the optimal step size and attack budget, which is given by
$\hat{\alpha} \approx 0.2 (\epsilon + 1)$.

Next, we investigate the effect of AT on the adversarial loss. \Figure{loss:eps8} shows the adversarial loss surface of the benign model for $\epsilon=8$ and \Figure{loss:eps8_at} shows the adversarial loss surface of an adversarially trained model 
for 10 epochs. We observe that, as the iterations proceed, (1) the adversarial loss surface becomes approximately piece-wise constant, (2) there is a wider range of values for the step size that yield an optimal attack, (3) the optimal value of the step size for the benign model is still optimal for the adversarially trained model, and (4) the adversarial loss value at the optimum decreases significantly. Therefore, for a fixed $\epsilon$, as the iterations of AT proceed, there is no need to change the value of $\alpha$ and finding effective attacks becomes harder.

\subsection{Adversarial Training with Optimal Attacks Gives Suboptimal Performance at Test Time}

Adversarial training is predicated on the idea of training with optimal attacks. For example, Huang \emph{et al.}~\cite{Huang2015LearningWA} claim that the key to solving the min-max problem of AT is to find optimal adversarial examples. The results from the previous section show us how to tune the parameters of PGD to obtain optimal attacks for video classification. In this subsection we wish to verify whether AT with optimal attacks gives a strong defense. Moreover, we wish to understand if the attack budget at training time should be chosen to be the same as the attack budget at test time.

We validate the effectiveness of AT with the PGD$_\infty$ attack for different attack budgets $\epsilon\in\{4,8,12\}$ and step sizes $\alpha \in \{\hat\alpha,\hat\alpha/2,\epsilon\}$ using the top-1 test accuracy on the UCF101 dataset. The results of the benign and adversarially trained models for different attack budgets at test time are shown in \Table{optim_vs_suboptim}. As expected, the benign model performs well on clean data ($\epsilon=0$), but its accuracy deteriorates quickly from 92.3\% to around 2\% when evaluated with adversarial examples. On the other hand, AT with optimal attacks, AT$(\epsilon=8,\alpha=\hat{\alpha})$, has lower standard accuracy (28.8\%) and higher robust accuracy (9.39\%-20.5\%) than the benign model, but it generally underperforms AT with sub-optimal attacks, AT$(\epsilon=8,\alpha\neq\hat\alpha)$, in terms of both standard and robust accuracy.

Another interesting observation is that the step size in sub-optimal regions controls the trade-off between standard accuracy and robustness. Loosely speaking, a smaller step size $(\alpha=\hat\alpha/2)$ helps preserve standard accuracy to some extent but suffers from poor robustness, while increasing the step size $(\alpha=\epsilon)$ improves robustness but reduces the standard accuracy. The results also show that using a stronger attack budget during training $(\epsilon=12)$ reduces both standard and robust accuracy, while using a weaker attack budget $(\epsilon=4)$ improves standard accuracy as well as robustness against weaker attack ($\epsilon=4)$.


Overall, the results thus far suggest that when the attack budget at test time is known, we should use AT with the same attack budget at training time and a suboptimal choice for the step size. However, when the attack budget at test time is unknown, we need a mechanism for selecting the attack budget and step size at training time. This motivates the extensions of AT that we discuss next.

\ifmain
\begin{table}
\centering
\caption{Top-1 test accuracy on the UCF101 dataset of both benign and adversarially trained models with PGD$_\infty$ for different attack budgets and step sizes. All attacks at test time are computed using PGD$_\infty$ with a fixed attack budget and the optimal step size.} 
\label{tab:optim_vs_suboptim}
\begin{adjustbox}{max width=\linewidth}
\begin{tabular}{llllll}
\toprule
\multicolumn{1}{c}{\multirow{2}{*}{Model}} & \multicolumn{5}{c}{Evaluation Attack} \\
                   & $\epsilon=0$ & $\epsilon=4$ & $\epsilon=8$ & $\epsilon=12$ & $\epsilon=15$ \\\midrule
Benign                               & \textbf{92.3}     & 2.65          & 2.38          & 2.11          & 1.89          \\
\midrule
AT$(\epsilon=8,\alpha=\hat{\alpha})$      & 28.8              & 20.5          & 13.9          & 10.6          & 9.39          \\
AT$(\epsilon=8,\alpha=\frac{\hat{\alpha}}{2})$ & \textit{72.9}              & \textit{40.0}          & 19.5          & 11.0          & 8.22          \\
AT$(\epsilon=8,\alpha=\epsilon)$       & 64.1              & {39.9} & \textbf{23.3} & \textbf{13.6} & \textit{10.6} \\
AT$(\epsilon=4,\alpha=\epsilon)$       & {70.1}     & \textbf{40.5} & \textit{19.9} & {12.8} & 10.4          \\
AT$(\epsilon=12,\alpha=\epsilon)$     & 34.3              & 24.8          & 17.8          & \textit{13.5} & \textbf{12.1} \\
\bottomrule
\end{tabular}
\end{adjustbox}
\end{table}
\fi
\ifworkshop
\begin{figure}[t]
\centering
\includegraphics[width=0.88\columnwidth]{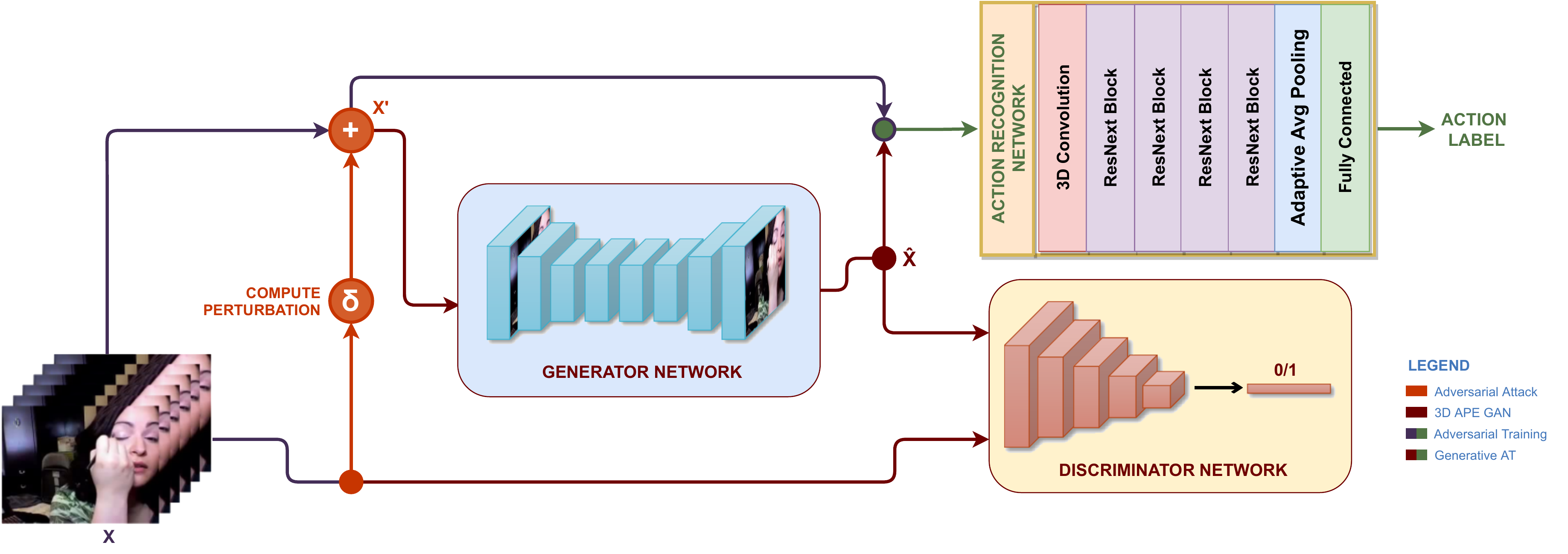}

\else
\begin{figure*}[h]
\centering
\includegraphics[width=1.9\columnwidth]{fig/AT_3D_APE.pdf}
\fi
\caption{
\textbf{Overall architecture of the proposed methods -- } Given a video $X$, an adversarial perturbation $\delta$ is added to it to generate an adversarial example $X'$, which is then passed to the action classifier model for Adversarial Training (AT). Adaptive AT and Curriculum AT extend AT by changing the attack budget and attack type during training. In the case of 3D APE-GAN, the generator network is trained to eliminate perturbations from $X'$ and reconstruct $\hat{X}$, which is then passed to the benign model, while the discriminator is trained to classify the original video $X$ from reconstructed video $\hat{X}$. Generative AT is an end-to-end integration of 3D APE-GAN and AT. }
\label{fig:3d_ape}
\ifworkshop
\end{figure}
\else 
\end{figure*}
\fi

\section{Extensions of Adversarial Training}
\label{sec:ATextensions}

Based on our findings from the previous section, in this section we propose three defenses against attacks with variable attack budgets. The first one, Adaptive AT, is a method where the attack budget is drawn from a distribution that is adapted as training iterations proceed. The second one, Curriculum AT, is a technique where the attack budget is increased as training iterations proceed. The third one, Generative AT, further couples AT with a denoising generative network to boost robust performance. The overall architecture of the proposed methods is illustrated in \Figure{3d_ape}.

\subsection{Adaptive Adversarial Training (AAT)}
\label{sec:AAT}
As it can be seen from the AT framework (see \Eq{at_basic}), the attack budget is predetermined for a fixed attack type. While this can provide robustness to that particular attack type and budget, it does not guarantee robustness against different attacks at test time~\cite{Tramr2019AdversarialTA}. For example, Schott \etal  \cite{Schott2019TowardsTF} show that increasing robustness to one attack type can decrease robustness to other attacks. Besides, the results in \Table{optim_vs_suboptim} show that a model trained with an attack budget $\epsilon$ almost always outperforms models trained with a different budget when evaluated on attacks with the same budget~$\epsilon$.

Here, we propose a technique to train the model with adversarial examples generated from a discrete set of attack types $\T$, \eg $\T=\{$Frame Saliency, Frame Border, PGD Patch, Flickering$\}$, or a discrete set of $B$ attack budgets $\E = \{\epsilon_1, \dots, \epsilon_B\}$ to yield simultaneous robustness against multiple attacks at a test time. A na\"ive approach to handle this could be to sample the attack type or budget from a uniform distribution on $\T$ or $\E$ and generate an attack with that type or budget. However, the resulting model could be more robust to some attacks and less robust to others. To address this issue, we propose to adapt the distribution of attack types or budgets depending on the performance of the current model for different attacks. For example, we can calculate the probability of attack budget $\epsilon_i \in \E$ as 
\begin{equation}
    \label{eq:adaptive}
    \mathbb{P}(\epsilon_i) \coloneqq \frac{\sum_{n=1}^N \mathcal{L}_{ce} (F(X_n+\delta_n^i;\theta_F), y_n)}{\sum_{j=1}^B \sum_{n=1}^N \mathcal{L}_{ce} (F(X_n+\delta_n^j;\theta_F), y_n)},
\end{equation}
where $\delta_n^i$ is an attack for $(X_n,y_n)$ with budget $\epsilon_i$. As adapting this probability at each epoch can be costly, we update it only every $\xi$ epochs. Algorithm~\ref{alg:adapt} summarizes the proposed AAT algorithm with adaptive sampling of multiple attack budgets. Multiple attack types can be handled similarly.

\setlength{\textfloatsep}{6pt}
\begin{algorithm}
\caption{Adversarial training with an adaptive random sampling of variable attack budget}
\label{alg:adapt}
\begin{algorithmic}[1]
\Require set of attack budgets $\E$, set of attack step sizes $\A$

	\Function{AdaptAT}{} 
	\State{randomly initialize $\mathbb{P}(\epsilon)$ for $\epsilon\in\E$} 
    \For{$k \gets 1$ to $K$} \Comment{number of epochs}
        \For{$n \gets 1$ to $N$} \Comment{number of batches}
        \State {$\epsilon \sim \mathbb{P}$} \Comment{sample attack budget}
        \State {$\alpha = \A(\epsilon)$} \Comment{set step size for $\epsilon$}
           
            \State $\delta_n \gets \Call{Attacker}{F,X_n,y_n,\epsilon,\alpha}$ \Comment{\Eq{pgd}}
            \State{$\theta_F \gets \theta_F - \eta  \nabla_{\theta_F} \mathcal{L}_{ce}(F(X_n+\delta_n;\theta_F),y_n)$}
        \EndFor
        \If{$k\mod{ \min(\xi, \frac{K}{|\E|})} \equiv 0$} 
            \For{$i \gets 1$ to $|\E|$}
                \State $\!\!\!\mathbb{P}(\epsilon_i) \coloneqq \frac{\sum_{n=1}^N \mathcal{L}_{ce} (F(X_n+\delta_n^i;\theta_F), y_n)}{\sum_{j=1}^B \sum_{n=1}^N \mathcal{L}_{ce} (F(X_n+\delta_n^j;\theta_F), y_n)}$
            \EndFor
        \EndIf 
    
    \EndFor
	\EndFunction
\end{algorithmic}
\end{algorithm}



\subsection{Curriculum Adversarial Training (CAT)} 
The framework of AT tries to optimize the inner maximization by generating the optimal adversarial examples, however, adversarially robust generalization might hurt the standard generalization~\cite{Tsipras2019RobustnessMB}. Plus, the optimal adversarial attack does not always lead to better defense, or it might even lead the training to fail and result in a worse generalization in both standard and robust performance~\cite{Zhang2020AttacksWD}. It has also been found that AT overfits the adversarial examples observed~\cite{Cai2018CurriculumAT,Rice2020OverfittingIA}. To mitigate all these issues, we propose a curriculum-based AT, wherein the model is trained with adversarial examples generated with an increasing attack budget at every $\xi^{th}$ iteration as the training proceeds. Cai \etal \cite{Cai2018CurriculumAT} propose a similar approach to the image classification task, however, it only increases the number of steps while the attack budget is fixed, and again this does not provide guaranteed robustness for other attack budgets~\cite{Tramr2019AdversarialTA}. 

\subsection{Generative Adversarial Training (GAT)}
Defense methods based on deep generative models such as GANs have shown promising results in the image domain. However, these techniques lag in video generation 
and state-of-the-art results are very far from satisfying~\cite{Clark2019AdversarialVG}. 

In this work, we propose 3D APE-GAN, an extension of APE-GAN~\cite{Jin2019APEGANAP} that removes perturbations from adversarial videos by projecting them onto a clean data manifold. Specifically, given an adversarial video, the 3D APE-GAN generator $G$ is trained to reconstruct a clean video, while the discriminator $D$ is trained to distinguish the original video from the video produced by the generator. 
This is achieved by solving a min-max problem of the form:
\begin{align}
\begin{split}
&\!\!\!\!
\min_{\theta_G}  \max_{\theta_D}
\mathop{\mathbb{E}}_{(X,y) \sim \mathcal{D}} 
\Big[\gamma_1 \mathcal{L}_{mse}(G(X+\hat\delta;\theta_G),X) \\
&\!\!\!\!+\gamma_2 \big (\log(D(X;\theta_D)) - \log(D(G(X+\hat\delta;\theta_G);\theta_D)) \Big], 
\end{split}
\end{align}
where $\theta_G$ denotes parameters of the generator, $\theta_D$ denotes parameters of the discriminator, $\mathcal{L}_{mse}$ is the MSE loss
 \begin{align}
     \mathcal{L}_{mse} &\coloneqq \frac{1}{T W H} \|X - G(X';\theta_G) \|_F^2,
 \end{align}
and $\hat\delta \coloneqq \argmax_{\delta \in \mathcal{B}_\epsilon} \mathcal{L}_{ce} (F(G(X+\delta;\theta_G);\theta_F), y)$. 

Notice that the min-max objective is a weighted combination of the MSE loss and the GAN objective with weights $\gamma_1$ and $\gamma_2$, respectively. We solve this problem in an alternating minimization fashion, where given $\theta_G$, we find~$\theta_D$~as
\begin{align}
\max_{\theta_D}
\mathop{\mathbb{E}}_{(X,y) \sim \mathcal{D}} 
\log(D(X;\theta_D)) - \log(D(G(X+\hat\delta;\theta_G);\theta_D).
\end{align}
Then, given $\theta_D$, we find $\theta_G$ as
\begin{align}
\begin{split}
\min_{\theta_G}
\mathop{\mathbb{E}}_{(X,y) \sim \mathcal{D}} 
\Big[\gamma_1 \mathcal{L}_{mse}(G(X+\hat\delta;\theta_G),X)\\
-\gamma_2 \log(D(G(X+\hat\delta;\theta_G);\theta_D)) \Big].
\end{split}
\end{align}
One extension we propose is to train 3D APE-GAN with AT so that the generator can be used as a preprocessing step to eliminate the perturbations before being fed to the video action recognizer. We can first train the 3D APE-GAN to generate clean videos given adversarial videos, freeze the pre-trained generator and adversarially train only the action recognition network $F$. 
\ifmain
This can be formulated as:
\begin{equation}
 \min_{\theta_F} \mathop{\mathbb{E}}_{(X,y) \sim \mathcal{D}} \left[ \max_{\delta \in \mathcal{B}_\epsilon} \mathcal{L}_{ce} (F(G(X+\delta;\theta_G);\theta_F), y) \right].\!\!
\end{equation}
\fi
Another extension we propose is to train 3D APE-GAN and the action recognition network end-to-end
\ifmain
 by optimizing:
\begin{align}
\label{eq:1}
\begin{split}
&\!\!\!\!\min_{\theta_F} \min_{\theta_G}  \max_{\theta_D}
\mathop{\mathbb{E}}_{(X,y) \sim \mathcal{D}} 
\Big[ \gamma_1 \mathcal{L}_{mse}(G(X+\hat\delta;\theta_G),X)  \\
&\!\!\!\!+ \gamma_2 \big (\log(D(X;\theta_D)) - \log(D(G(X+\hat\delta;\theta_G);\theta_D)\big)\\
&\!\!\!\!+\mathcal{L}_{ce} (F(G(X+\hat\delta;\theta_G);\theta_F), y) \Big]. 
\end{split}
\end{align}
\fi

\section{Experiments and Results}
As discussed before, the results in \Table{optim_vs_suboptim} show that a model trained with an attack budget $\epsilon$ almost always outperforms models trained with a different budget when evaluated on attacks with the same budget~$\epsilon$. This suggests we need to train our model with multiple attack budgets to provide better robustness to different attack at test time. In this section, we evaluate the performance of the proposed AAT, CAT and GAT methods against multiple attack types.

\subsection{AT with Variable Attack Budget}
The first method we evaluate is AAT, \ie, the proposed adversarial training method with an adaptive random sampling of variable attack budgets, as presented in Algorithm~\ref{alg:adapt}. The network is trained with a set of attack budgets $\E = \{0,2,4,..,12\}$ sampled based on the probability distribution computed via~\Eq{adaptive} updated every $\xi=10$ epochs. The second approach we evaluated is the proposed Curriculum AT (CAT) where we increase the attack budget from $\epsilon=0$ to $\epsilon=12$ or decrease it from $\epsilon=12$ to $\epsilon=0$ as the training proceeds. The overall results are shown in \Table{variable_budget}.

\ifmain
\begin{table}
\centering
\caption{
\textbf{Adversarial training with variable attack budgets} -- Top-1 test accuracy on the UCF101 dataset of both benign and adversarially trained models with PGD$_\infty$ for variable attack budgets and step sizes. All attacks at test time are computed using PGD$_\infty$ with a fixed attack budget and the optimal step size. $AAT$ with attack budget $\epsilon=auto$ represents the adaptive random sampling method of variable attack budgets. $CAT$ with $\epsilon=\uparrow_0^{12}$ indicates curriculum training with an increasing attack budget from $\epsilon=0$ to $\epsilon=12$, while CAT with $\epsilon=\downarrow_0^{12}$ indicates a decreasing budget from $\epsilon=12$ to $\epsilon=0$.  
} 
\label{tab:variable_budget}
\begin{adjustbox}{max width=\linewidth}

\begin{tabular}{llllll}
\toprule
\multicolumn{1}{c}{\multirow{2}{*}{Model}} & \multicolumn{5}{c}{Evaluation Attack} \\
                   & $\epsilon=0$ & $\epsilon=4$ & $\epsilon=8$ & $\epsilon=12$ & $\epsilon=15$ \\\midrule
\midrule
Benign                                                   & \textbf{92.3}     & 2.65          & 2.38          & 2.11          & 1.89   \\
AT$(\epsilon=8,\alpha=8$)                           & \textit{64.1}     & \textit{39.9} & {23.3}        & {13.6}        & {10.6} \\
\midrule
AAT$(\epsilon=auto,\alpha=\epsilon)$           & {54.5}            & {38.3}        & \textit{25.7} & \textit{18.7}   & \textit{16.0} \\
CAT$(\epsilon=\downarrow_0^{12},\alpha=\epsilon)$ & {47.8}        & {24.4}        & {11.5}        & 6.61          & {5.40}        \\
CAT$(\epsilon=\uparrow_0^{12},\alpha=\epsilon)$ & 61.1            & \textbf{42.2} & \textbf{29.2} & \textbf{21.0} & \textbf{17.2} \\

\bottomrule
\end{tabular}
\end{adjustbox}
\end{table}

\begin{table}[b]
\centering
\caption{
\textbf{3D APE-GAN with AT} -- Top-1 test accuracy on the UCF101 dataset of both benign and adversarially trained models with PGD$_\infty$ for different attack budgets and step sizes. All attacks at test time are computed using PGD$_\infty$ with a fixed attack budget and the optimal step size. APE denotes the proposed 3D APE-GAN method using benign action recognition network, while GAT represents the Generative AT, i.e. end-to-end adversarial training of both 3D APE-GAN and action recognition network.
} 
\label{tab:3d_ape_gan}
\begin{adjustbox}{max width=\linewidth}
\begin{tabular}{@{}llllll@{}}
\toprule
\multicolumn{1}{c}{\multirow{2}{*}{Model}} & \multicolumn{5}{c}{Evaluation Attack} \\
                   & $\epsilon=0$ & $\epsilon=4$ & $\epsilon=8$ & $\epsilon=12$ & $\epsilon=15$ \\\midrule
\midrule
Benign                               & \textbf{92.3}     & 2.65         & 2.38              & 2.11          & 1.89          \\
AT$(\epsilon=8,\alpha=8$)               & {64.1}             & {39.9}       & {23.3}            & {13.6}        & {10.6}        \\
CAT$(\epsilon=\uparrow_0^{12},\alpha=\epsilon)$ & 61.1          & \textbf{42.2} & \textbf{29.2} & \textbf{21.0} & \textbf{17.2} \\
\midrule
APE$(\epsilon=\uparrow_0^{12},\alpha=\epsilon)$   & \textit{81.6} & \textit{69.7} & \textit{59.8} & \textit{54.6} & \textbf{54.1} \\
GAT$(\epsilon=8,\alpha=8$)                   & 81.1          & \textbf{72.6} & \textbf{63.8} & \textbf{56.6} & \textit{51.3}  \\
\bottomrule
\end{tabular}
\end{adjustbox}
\end{table}
\fi

\ifmain
\begin{table*}[t]
\centering
\caption{
\textbf{Robustness to multiple attack types} --Test accuracy on 100 videos of the UCF-101 dataset of benign and adversarially training models with different video attack types. For each attack type, the results presented here are an average top-1 test accuracy of three experiments with varying parameters. The results of each attacks with the different parameters is depicted in \Figure{multiple_attack}.
} 
\label{tab:multiple_attack}
\begin{adjustbox}{max width=0.99\linewidth}
\begin{tabular}{@{}lcccccc@{}}
\toprule
\multicolumn{1}{c}{\multirow{2}{*}{\begin{tabular}[c]{@{}c@{}}\\Model\end{tabular}}} & \multicolumn{6}{c}{Evaluation Attack} \\
\multicolumn{1}{c}{}                     & \multicolumn{1}{c}{No attack} & \multicolumn{1}{c}{\begin{tabular}[c]{@{}c@{}}Frame saliency\\(one-shot)\end{tabular}} & \multicolumn{1}{c}{\begin{tabular}[c]{@{}c@{}}Frame saliency\\(iterative)\end{tabular}} & \multicolumn{1}{c}{Masked PGD} & \multicolumn{1}{c}{Frame Border} & \multicolumn{1}{c}{Flickering} \\
\midrule \midrule
Benign                           & \textbf{93.0}  &   0.00	&	0.00	&	7.0	&	0.00	&	42.3         \\
Video Compression~\cite{art2018}              & 92.6	&	48.6	&	50.0	&	7.0	&	10.0	&	40.3 \\
\midrule
AAT $(\epsilon=auto,\alpha=\epsilon)$                 & 92.6	        &	\textit{74.4}	&	{74.9}     	&	\textit{67.5}  &	\textit{80.2}	&	{53.1}       \\
APE $(\epsilon=auto,\alpha=\epsilon)$        & \textit{92.7}	&	74.0	        &	\textit{76.6}  &	63.1	        &	63.4	        &	\textit{55.3} \\
GAT $(\epsilon=auto,\alpha=\epsilon)$   & 79.3	        &	\textbf{85.1}	&	\textbf{86.3}	&	\textbf{81.9}	&	\textbf{86.6}	&	\textbf{58.1} \\
\bottomrule
\end{tabular}
\end{adjustbox}
\end{table*}
\fi

As we can see from the results, AAT improves the robust performance (especially against stronger attacks). However, the standard accuracy reduces by almost 10\% when compared to AT. On the other hand, CAT with an increasing attack budget outperforms all the other previous methods in all adversarial attacks while it lessens the standard accuracy by only 3\% from AT. 
\ifmain
That being said, although CAT improves robust accuracy, the overfitting problem~\cite{Rice2020OverfittingIA} is still not fully solved: \Figure{cat_gen_gap} shows that the generalization gap goes up as the training iterations and attack budget simultaneously increase. This demonstrates that the generalization gap when training with stronger attacks is higher than with weaker attacks. 
Thus, this suggests that further work is needed to close this gap in order to yield better robustness against optimal attacks at test time.
\begin{figure}[h]
\centering
\includegraphics[width=0.9\linewidth,clip=true,trim=0 15 15 75]{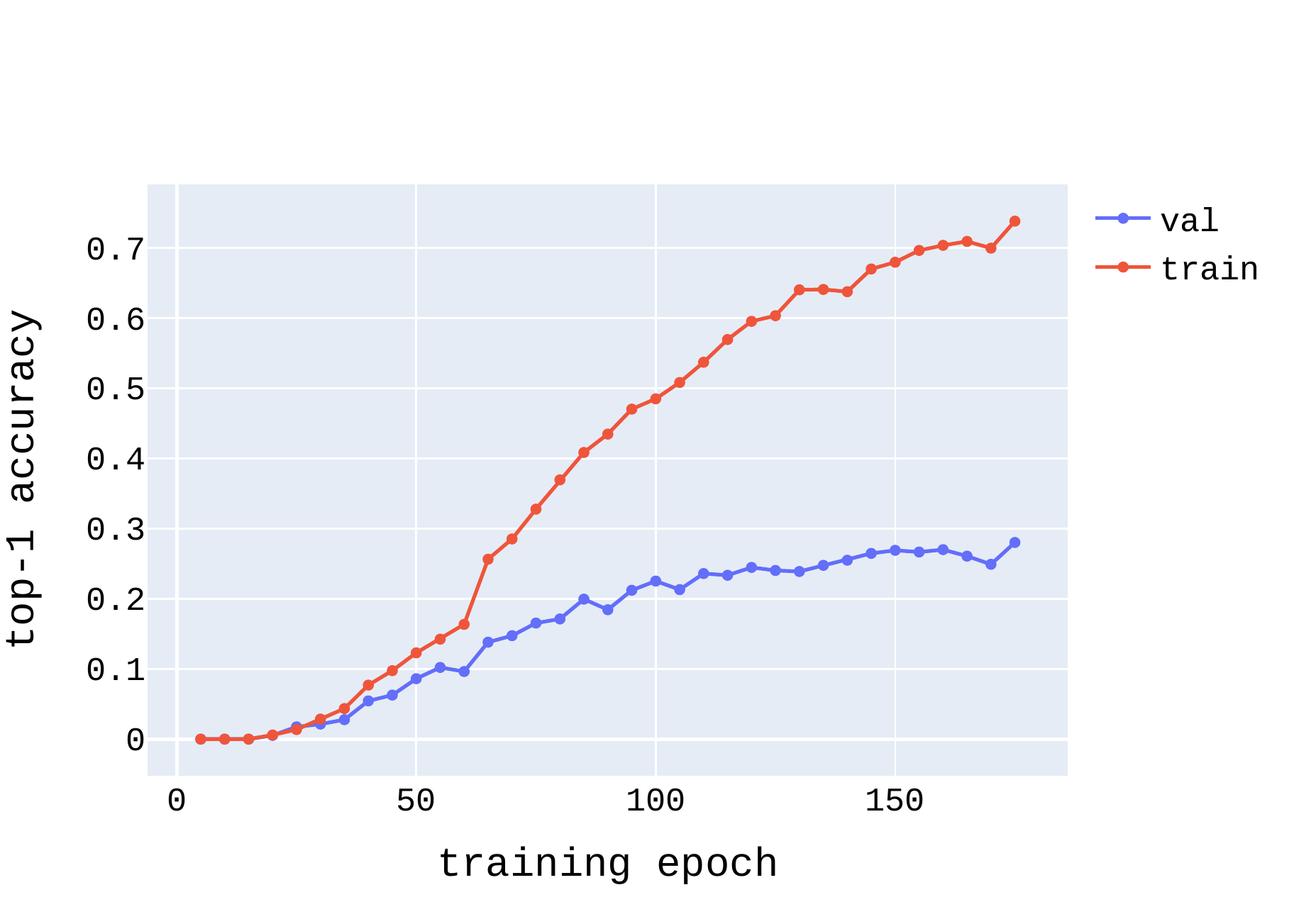}
\caption{
\textbf{Generalization gap of curriculum adversarial training  -- }
The generalization gap keeps increasing as the curriculum training proceeds. The $x$-axis is training iterations and the $y$-axis is the top-1 accuracy. 
The red and blue lines indicate the top-1 training and validation accuracy, respectively.
}
\label{fig:cat_gen_gap}
\end{figure}

\fi 
\begin{figure}
\centering
\includegraphics[width=0.9\linewidth,clip=true,trim=20 0 0 0]{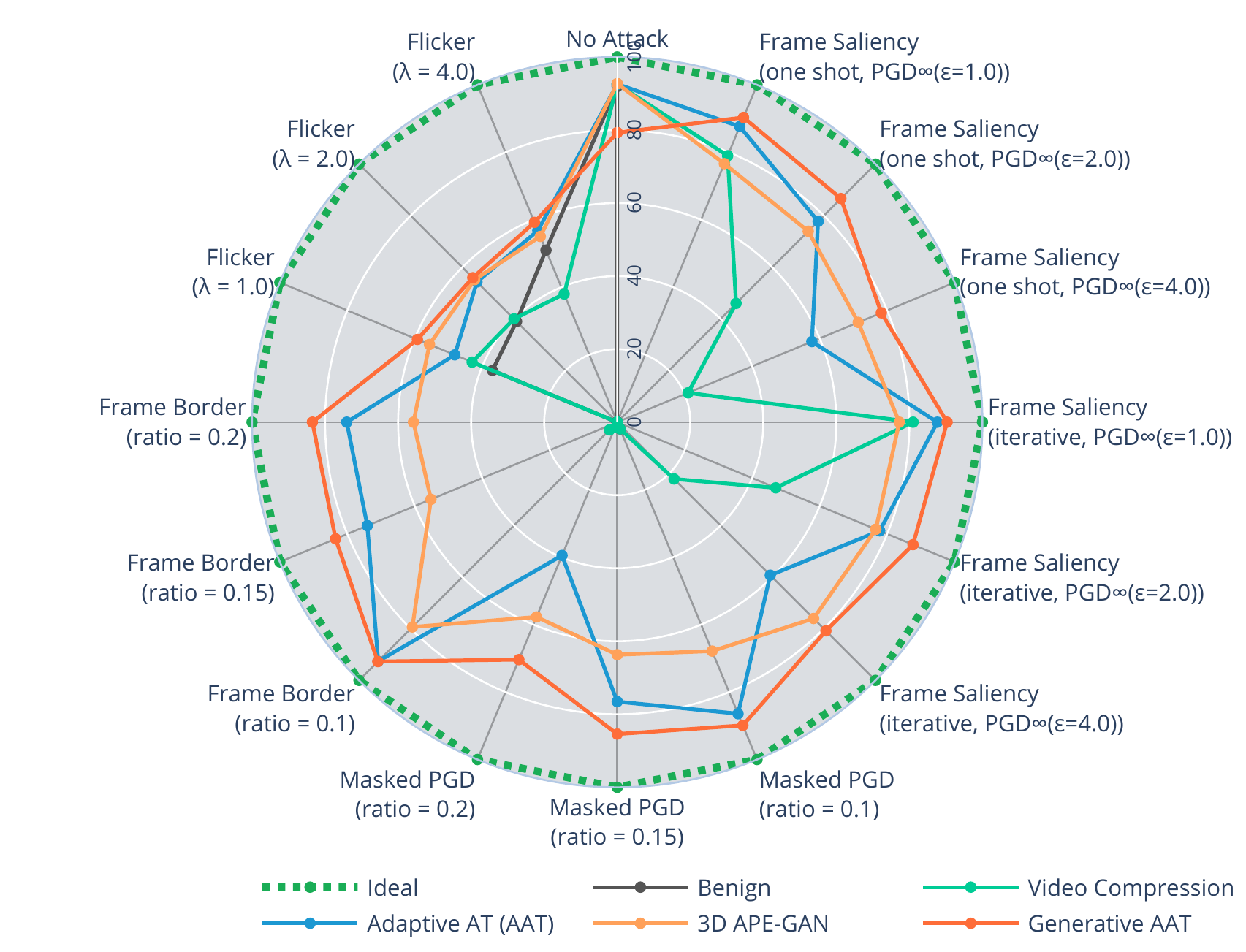}
\caption{
\textbf{Robustness to multiple attack types} -- robust performance of models trained with different video attack types evaluated on 100 videos of the UCF-101 dataset. 
}
\label{fig:multiple_attack}
\end{figure}

\subsection{Generative Adversarial Training}
In this experiment, we evaluate the 3D APE-GAN method and the end-to-end trainable GAT method.
\ifmain
The performance of both techniques with the previous best performing technique is presented in \Table{3d_ape_gan}. 
\fi
The results show that the 3D APE-GAN with benign model significantly improves the standard and robust accuracy when compared with previous AT methods. Moreover, GAT further improves the performance in almost all evaluations. However, note that the generator is being used as a pre-processing step and thus it is not being attacked.

\subsection{AT with Multiple Attack Types}
In this experiment, we evaluate the effectiveness of our methods against different attack types. We consider five video attack methods -- frame saliency (one-shot), frame saliency (iterative), masked PGD, frame border, and flickering attacks. 
The baseline method we use is the video compression defense~\cite{art2018}. The methods are trained with adversarial examples generated from an adaptive distribution of four attack methods (except the flickering attack). The robustness performance of the methods to the different video attacks is illustrated in \Figure{multiple_attack}. 
\ifmain
The average performance against each attack type is listed in \Table{multiple_attack}. \fi
From the results, it is evident that AAT provides robustness to multiple attack types and Generative Adaptive AT boosts the robustness, but reduces the standard accuracy by just above 10\%.

\section{Conclusions}
In this paper, we showed that generating optimal attacks for video requires carefully tuning the attack parameters, especially the step size. We also showed that the optimal step size varies linearly with the attack budget. We demonstrated that using a sub-optimal attack budget at training time leads to more robust performance at testing, and that the generalization gap rises as the attack budget increases during testing. Based on these findings, we proposed three defenses against attacks. Adaptive AT and Curriculum AT extend AT by using variable attack budgets as the training iterations proceed. This is achieved by adapting the distribution of attack budgets to the performance and increasing the attack budget, respectively. Generative AT integrates AT with a denoising generative network to further improve robust performance. Our proposed approaches significantly increase the robustness to multiple attack budgets and attack types. 

\myparagraph{Acknowledgments}
The authors thank Ambar Pal 
for 
his 
valuable comments and contribution to the development of this work. This research work was supported by DARPA GARD Program
HR001119S0026-GARD-FP-052.

{
    \small
    \bibliographystyle{ieee_fullname}
    \bibliography{macros,cvpr22-AT}
}



\end{document}